# STEM - CALYX RECOGNITION OF AN APPLE USING SHAPE DESCRIPTORS

Mohana S.H., Prabhakar C.J

Department of P.G. Studies and Research in Computer Science, Kuvempu University
Shankaraghatta-577451, Shimoga, Karnataka, India

## ABSTRACT

*This paper presents a novel method to recognize stem - calyx of an apple using shape descriptors. The main drawback of existing apple grading techniques is that stem - calyx part of an apple is treated as defects, this leads to poor grading of apples. In order to overcome this drawback, we proposed an approach to recognize stem-calyx and differentiated from true defects based on shape features. Our method comprises of steps such as segmentation of apple using grow-cut method, candidate objects such as stem-calyx and small defects are detected using multi-threshold segmentation. The shape features are extracted from detected objects using Multifractal, Fourier and Radon descriptor and finally stem-calyx regions are recognized and differentiated from true defects using SVM classifier. The proposed algorithm is evaluated using experiments conducted on apple image dataset and results exhibit considerable improvement in recognition of stem-calyx region compared to other techniques.*

## KEYWORDS

*Stem-calyx recognition, Grow-cut, Multithreshold segmentation, Multifractal dimension, Radon transform, Fourier transform, SVM*

## 1. INTRODUCTION

Surface defect detection for inspection of quality of apples has been gaining importance in the area of machine vision. Color, size and amount of defects are important aspects for inspection and grading of fresh apples. Surface defects are of great concern to farmers and vendors to grade the apples. The defect or damage is usually occurred in apples due to various factors. Apples with rot, bruising, scab, fungal growth, injury, disease and other defects must be removed to prevent cross-contamination and reduce subsequent processing cost. There are two important stages involved in grading the apples such as surface defect detection and grading the apples based on identified defects. The researchers have adopted image processing and computer vision based techniques to detect the surface defects of apples, and apples are graded based on quantity of surface defects detected using either RGB images (B.S Bennedsen et al., 2005; Prabhakar C.J et al., 2013) or multispectral images (J.A Throop et al., 2005; V Leemans et al., 2002; Xianfeng Li et al., 2011).
The main drawback of these techniques is that while identifying the defects, stem-calyxes are identified as defects because after applying image processing techniques, stem and calyx region appears to be similar to defect spots. Due to similarity in appearance between stem-calyx and defects, the defect detection algorithms wrongly identify the stem-calyx parts as defects, which reduce the grading accuracy. Some researchers have proposed techniques to differentiate actual defects and stem-calyx, but these techniques are not able to give the promising results. In this paper, we proposed a technique to recognize stem-calyx and differentiated from true defects using

RGB image of an apple. The motivation to use RGB images is that unlike the MIR camera, digital cameras are inexpensive and easy to afford. Another advantage of digital data is that they can be readily processed using digital computers.

The techniques for stem-calyx recognition of apples can be classified into two categories. The first category of techniques differentiates stem-calyx from true defects based on the threshold employed on segmented apple image. Whereas the second categories of techniques make the system to learn shape of stem-calyx part and true defects by extracting shape features and recognizes stem-calyx part using classifier.

Threshold-Based Techniques

The X.Cheng et al. (2003) have proposed an approach for stem-calyx recognition in multispectral images of apples. Brightness effect is eliminated using normalization and adaptive spherical transform is used to distinguish the low- intensity defective portions from the actual stem-calyx. The global threshold is used to extract the stem-calyx. V. Leemans et al. (2002) developed apple grading technique according to their external quality using machine vision. Color grading is done by Fisher's linear discriminant analysis and stem-calyx is detected by using correlation pattern recognition technique. The main drawback of the threshold based segmentation is that misclassification may occur if the threshold of stem-calyx and true defects are similar.

Features-Based Techniques

D.Unay et al. (2007) have proposed an approach to recognize stem-calyx of apples using monochrome images. Fruit area is separated from background using threshold based segmentation. Object segmentation technique is used for stem-calyx detection and it is done by thresholding. Statistical, shape and texture features are extracted. They conclude that SVM classifier gives best performance in recognizing stem-calyx. J Xing et al. (2007) proposed an approach for stem-calyx identification for multispectral images. In order to identify the stem/calyx on apples, the technique is developed based on contour features in the PCA image scores. The main drawback of these techniques is that they used MIR camera to capture multispectral images, which are too expensive to use, and it is very difficult to process the huge amount of data.

Dong Zhang et al. (2013) developed an automated apple stem-calyx detection using evolution constructed features for Gala apples. Series of transforms were employed and AdaBoost classifier is used to differentiate stem-calyx from true defects. L Jiang et al. (2009) developed a 3D surface reconstruction of apple image and analyzed the depth information of apple surface to identify stem-calyx part. Qingzhong Li et al. (2002) proposed an approach to detect the defects in the surface of an apple. Stem-calyx part is identified using fractal features, which are fed into ANN classifier. They considered mono fractal features to classify the stem-calyx part assuming that concave surface. However, this is not true in case where apples are oriented along in any direction and the concave surface may not exist. Another drawback of this technique is that they used fractal features alone to distinguish the stem-calyx from true defects.

In order to overcome the drawback of mono fractal features where these features failed to represent the stem-calyx part exists in the apple image with non-concave surface. We proposed an approach to recognize the stem-calyx based on multifractal features. The recognition accuracy of our technique is increased by combining multifractal descriptor with other two shape descriptors such as Fourier and Radon transform based descriptors.

The organization of remaining sections of the paper is as follows: In section 2, theoretical description of proposed shape descriptors are given. The proposed approach for stem-calyx recognition is discussed in the section 3. The experimental results are demonstrated in Section 4. Finally, Section 5 draws the conclusion.

## 2. SHAPE DESCRIPTORS

In many applications, like stem-calyx recognition, the internal content of the shape is not an important compared to the boundary of an object. The boundary-based techniques tend to be more efficient for handling shapes that are describable by their object contours. The simplified representation of an object is often called the shape descriptors or signatures. These simplified representations are easier to store, compare and handle. The descriptors for irregular shapes like stem-calyx should be different enough that the shapes can be discriminated.

### 2.1. Multifractal Descriptor

The fractal dimension is a very popular concept in mathematics due to its wide range of applications. The fractal dimension was developed by Benoit Mandelbrot in 1967 (Benoit M, 1967). Fractals are generally self-similar and independent of scale. A fractal is a fragmented geometrical object that can be divided into parts, and each part is a copy of the original in reduced size. The important characteristic of fractal geometry is that ability to represent an irregular or fragmented shape of complex objects, where as traditional Euclidean geometry fails to analyze. A multifractal is a set of intertwined fractals. A multifractal object is more complex in the sense that it is always invariant by translation. The multifractal dimensions were defined based on partition function.

In our approach, we used Box Counting method to compute the multifractal dimension. Consider the mass dimension α at a point x, the box $B_\varepsilon(x)$ is a box of radius ε centered at x. The probability measure of mass in the box is defined as $\mu(B_\varepsilon(x))$. It can be shown that $\mu(B_\varepsilon(x)) \propto \varepsilon^\alpha$. The mass dimension specifies how fast the mass in the box $B_\varepsilon(x)$ decreases as the radius approaches to zero. In the first step, we compute the total number of boxes required to cover the input image. The total number of boxes can be calculated using below equation

$$N_T = \frac{(width)^2}{(\varepsilon)^2}, \quad (1)$$

where, width size of the input image and ε is size of the box. It is necessary to eliminate the boxes which are having zero elements. Therefore, we select the boxes with non-zero elements and are defined as

$$N_S = \sum_{i=1}^{N_T} N_i = not\ null, \quad (2)$$

where $N_S$ represents total number of boxes with non-zero elements. The total sum or mass of pixels in all boxes can be obtained using the following equation

$$M_\varepsilon = \sum_{i=1}^{N_S} m_{[i,\varepsilon]}, \quad (3)$$

where $m_{[i,\varepsilon]}$ represents the mass of ith box with size ε. The probability P of the mass at ith box, is relative to the total mass $M_\varepsilon$ and is defined as

$$P_{[i,\varepsilon]} = \frac{m_{[i,\varepsilon]}}{M_\varepsilon}, \quad (4)$$

Signal & Image Processing : An International Journal (SIPIJ) Vol.5, No.6, December 2014

the qth order normalized probability measures of a variable, µ(q , ε ) vary with the box size ε and it is defined as

$$\mu_i(q,\varepsilon) = \frac{[p_i(\varepsilon)]^q}{\sum_i [p_i(\varepsilon)]^q}, \qquad (5)$$

where $p_i(\varepsilon)$ is the probability of a measure in the ith box of size ε. It can be shown that $P_{[i,\varepsilon]} \propto \varepsilon^\alpha$, the exponent α is called the Holder exponent at the point x, which is defined as

$$\alpha(q,\varepsilon) = \sum_i \mu_i(q,\varepsilon)\log p_i(\varepsilon), \quad \text{for } i=1,2,3,\ldots,N_S. \qquad (6)$$

For each q value, multifractal dimension is calculated as

$$f(q,\varepsilon) = \sum_i \mu_i(q,\varepsilon)\log \mu_i(q,\varepsilon), \quad \text{for } i=1,2,3,\ldots,N_S. \qquad (7)$$

## 2.2. Fourier Descriptor

Fourier descriptor is commonly used in popular areas such as pattern recognition, computer vision and image analysis. Fourier transformation on shape signatures is widely used for shape analysis, and also some researchers are used for shape retrieval. In the Fourier descriptor (Charles T Zhan et al., 1972), Fourier transformation and Fourier inverse transformation pairs were used to extract shape features from candidate objects. In general, Fourier descriptor (FD) is obtained by applying Fourier transform on a shape signature that is a one-dimensional function which is derived from the shape boundary coordinates. The normalized Fourier transformed coefficients are called the Fourier descriptor of the shape in that sample image. The boundary of an object is plotted on the XY plane. Starting at a arbitrary point (x0 , y0), coordinate pairs (x0 , y0), (x1 , y1), (x2 , y2), . . . , (xN-1 , yN-1) traversing the boundary. Each coordinate pair can be treated as a complex number, it can be represented as s(k) = x(k) + iy(k). The x-axis is treated as the real axis and the y axis is treated as the imaginary axis of a sequence of complex numbers. The advantage of this representation is that it reduces a 2-D problem into 1-D problem.

The Discrete Fourier Transform (DFT) of $s(k)$ is

$$a(u) = \frac{1}{N}\sum_{k=0}^{N-1} s(k)\, e^{-j2\pi uk/N}, \quad \text{for } u=0,1,2,\ldots,N\text{-}1. \qquad (8)$$

The complex coefficients a(u) are called the Fourier Descriptors of the boundary. The Fourier Descriptors are a frequency based description of the boundary of an image. The inverse Fourier transform of $a(u)$ restores $s(k)$. That is,

$$s(k) = \sum_{u=0}^{N-1} a(u)\, e^{j2\pi uk/N}, \quad \text{for } k=0,1,2,\ldots,N\text{-}1. \qquad (9)$$

## 2.3. Radon Descriptor

The Radon transform (RT) was introduced by the Austrian mathematician Johann Radon in 1917. The integral transform is consisting of the integral of a function over straight lines. The Radon Transformation is a fundamental tool which is used in different applications such as radar imaging, geophysical imaging and medical imaging. Sambhunath Biswas et al. (2012) developed an efficient face recognition algorithm based on transformed shape features. By definition the Radon transform of an image is determined by a set of projections of the image along lines taken at different angles. For discrete binary image data, each non-zero image point is projected into a Radon matrix. Let f(x, y) be an image. Its Radon transform is defined by

$$R(\rho,\theta) = \int_{-\infty}^{+\infty}\int_{-\infty}^{+\infty} f(x,y)\,\delta\,(\rho - x\cos\theta - y\sin\theta)dxdy, \qquad (10)$$

where $\delta(.)$ is the Dirac function, $\theta \in [0, \pi]$ and $\rho \in [-\infty, +\infty]$. The Radon transform has several useful properties. Some of them are relevant for shape representation: A translation of f result in the shift of its transform in the variable ρ by a distance equal to the projection of the translation vector on the line $\rho = x \cos\theta + y \sin\theta$. A rotation of the image by an angle $\theta_0$ implies a shift of the RT in variable θ. The Radon Transform is robust to noise, provided with fast algorithms, and it projects a two-dimensional function into one-dimensional function.

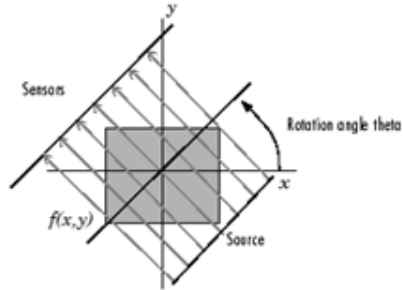

Figure 1. Parallel–beam projection at rotation angle theta

Radon transform is based on the parameterization of straight lines and the evaluation of integrals of an image along these lines. To represent an image, the radon function takes more than one parallel-beam projections of the image from different angles by rotating the source around the center of the image. The line integral of f(x,y) in the vertical direction is the projection of f(x,y) onto the x-axis. The line integral in the horizontal direction is the projection of f (x,y) onto the y-axis.

## 3. OUR APPROACH

In the training phase, we detect candidate objects such as stem part, calyx part and defected part from training samples and shape features are extracted from the detected candidate objects. As we described in the previous section, we extract and store the description of shape features using Multifractal, Fourier and Radon descriptors. The feature extraction step for each candidate object yields three feature vectors. The extracted feature vectors from stem part, calyx part and defected part of various training samples captured in different viewpoint are stored independently in the database and these stored feature vectors are treated as reference in order to recognize the stem-calyx part. In the testing phase, we recognize the stem-calyx by extracting the feature vectors from detected candidate objects of testing sample and compared with stored feature vectors using classifier.

### 3.1 Candidate Objects Detection

In order to detect the candidate objects such as stem part, calyx part and defected part in an apple image, we used the following steps. Pre-processing using a median filter, grow-cut approach for background subtraction, it is followed by multi-threshold segmentation approach to segment the candidate objects. Finally, the results of grow-cut and multi-threshold segmentations are combined to detect the candidate objects.

*A.* Image pre-processing

In order to improve the quality of an image, operations need to be performed on it to remove or decrease degradations suffered by the image during its acquisition. The captured apple images are suffered from illumination variations due to specular reflection. The specular reflection leads to poor performance during candidate objects detection process because the specular reflected part is treated as the candidate object due to its white color compared to its surroundings.

We employ median filtering to normalize the uneven distribution of light and to suppress noise. The median is much less sensitive than the mean to extreme values, i.e., they are the outliers. So, here median filtering is used to remove these outliers without reducing the sharpness of the image. The median filter technique allowed the edges to be preserved while filtering out the peak noise. For this reason, the median filter is often used before applying candidate object detection technique to preserve the shape of the candidate objects as much as possible.

*B.* Background Subtraction

The apple images were captured on a uniform color background. It is necessary to extract the ROI i.e. fruit area from the apple image. We used grow-cut approach for background subtraction. It removes background and gives fruit part separately. Vladimir Vezhnevets et al. (2005) from the Media Laboratory at Moscow State University first came up with this algorithm. The unique feature of the grow cut algorithm is in its ability to incorporate both "Von Neumann neighborhood" theory and the "region growing/region merging" approach.

Multi-Threshold Segmentation

After extracting the apple fruit part from the image, we segment the candidate objects from fruit area using multi-threshold segmentation. In multi-threshold segmentation, the images were segmented several times at different threshold levels. We experimentally observed that the candidate objects of various types of apples can be segmented accurately using threshold values 30, 50 and 65. The resulting binary images were added to form a so called multi layer image. This in turn was then subjected to threshold segmentation. This segmentation aimed at identifying and separates the healthy skin of an apple from the stem-calyx and defected areas in the original image. The resulting binary image was referred to as a marker image. The final step consists of constructing a binary image based on the marker image and the multi-layer image. With the position of the candidate objects, a simple thresholding routine i.e. a gradient segmentation is employed to determine the area of these candidate objects. The results obtained from the background subtraction and multi-threshold segmentation methods are overlapped to each other. In the resultant image, we observe that the candidate objects are appeared on the surface of the apples.

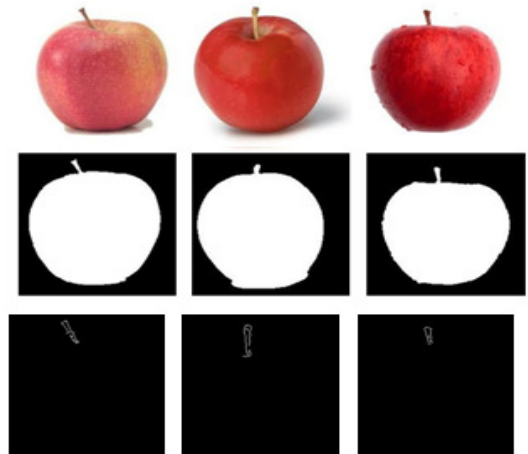

Figure 2: Candidate object detection process: First row: The pre-processed Jonagold apple images with stem part in view, Second row: fruit area extracted from images shown in first row, Third row: Detected stem part using multi-threshold segmentation.

## 3.2 Shape Features Extraction from Candidate Objects

In order to recognize stem-calyx part, we extract shape features because of its discriminative characteristics to distinguish stem-calyx part from defects. We extract the shape features from detected stem-calyx part and defect parts of training sample where the stem-calyx part and defect parts are captured in different views with variation in scale and views. The extracted shape features represent stem-calyx part and defect parts and invariant to variation in views and scale. We used three shape descriptors such as Multi-fractal descriptor, Fourier descriptor and Radon descriptor to extract and store the shape features independently for each detected candidate object.

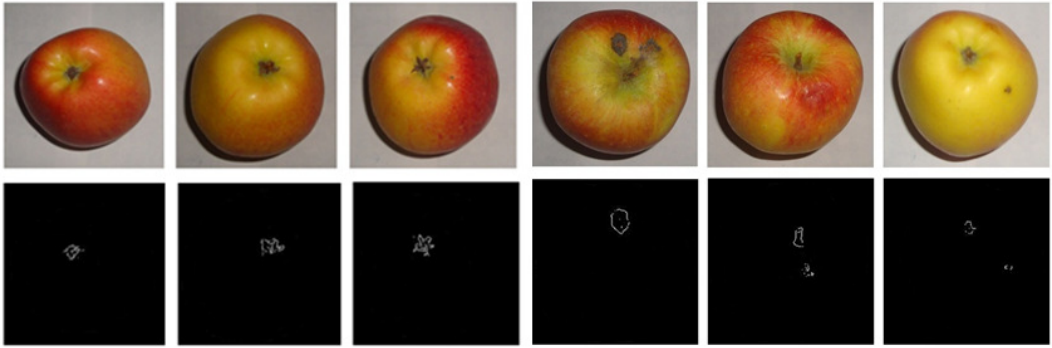

Figure 3: First row: the apple images with stem/calyx and defected part in view. Second row: detected stem-calyx part with defects.

Multifractal Descriptor

We extract multifractal dimension features of detected candidate objects using box counting approach with the box size 2 x 2 (4 pixels). Here we varied the q value from -1 to +1 (21 points). After fitting the boxes to the candidate image, we select the boxes which are having non-zero pixel (mass distribution) in the box, and the empty boxes are not considered for further processing. We calculate the mass distribution of all selected boxes and the probability of each box. The exponent **α** is called the Holder exponent is calculated using the probability measure. Finally, Multi-fractal dimension is calculated using normalized probability. In the Figure 4, from the plotted multifractal dimension graph for stem-calyx part and defected part, it is observed that the multifractal dimension exhibit discrimination power for stem, calyx and defected part.

Fourier Descriptor

The Fourier descriptors of the detected candidate objects are formed by Fourier transformed coefficients. The general features of the candidate object shape are contained in the lower frequency descriptors and finer details of the candidate shape are contained in the higher-frequency descriptor. High-frequency information is ignored because it is not helpful in shape discrimination. The complete numbers of coefficients are generated from the transform is very large, a subset of the total coefficients is enough to capture the overall features of the shape. The Fourier descriptor is clear representation of detected candidate objects, which can be observed in the Figure 5, where the reconstructed candidate object using the small set of Fourier coefficients appear to be as like original candidate object.

Signal & Image Processing : An International Journal (SIPIJ) Vol.5, No.6, December 2014

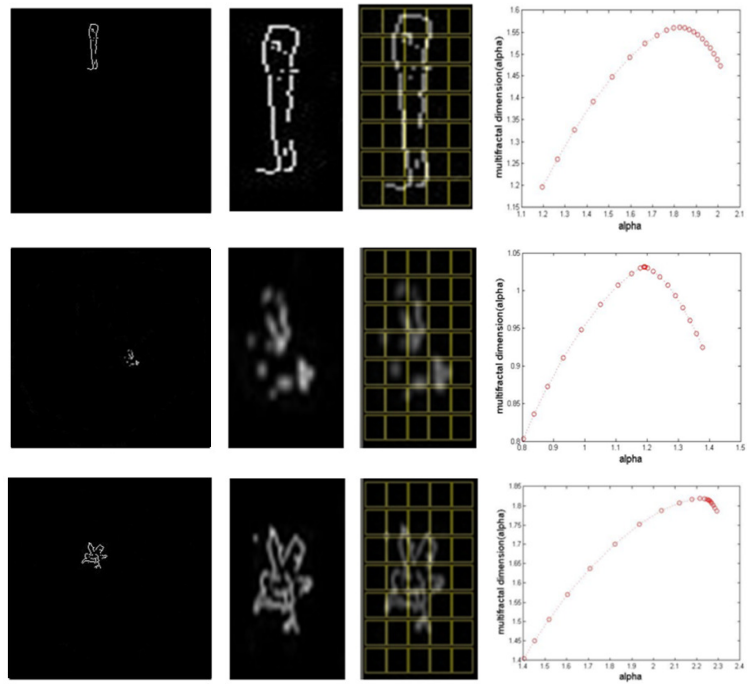

Figure 4: Results of multifractal descriptors: First column- Detected candidate objects, Second column- Cropped image of candidate object, Third column-Schematic diagram of box counting method, Last column- Corresponding graph of multifractal dimension.

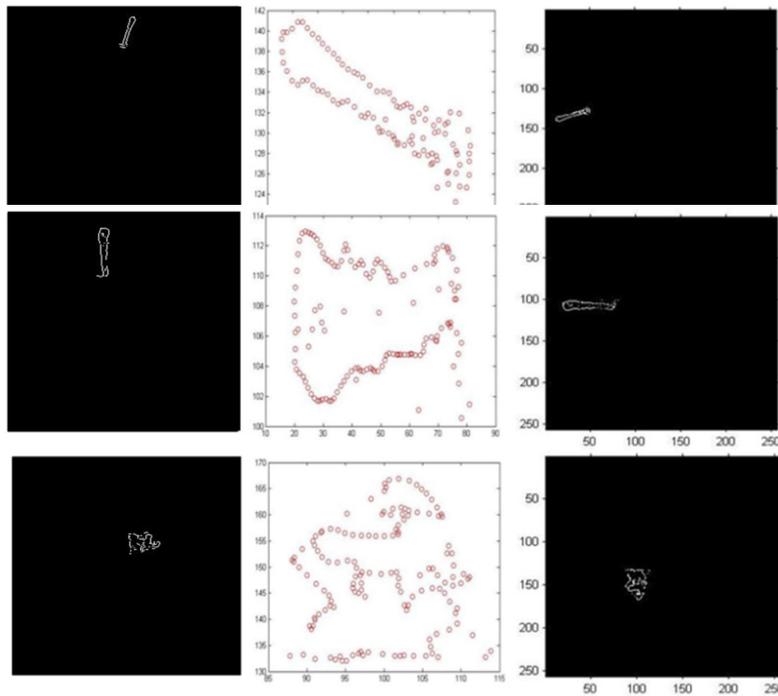

Figure 5: Results of Fourier descriptor: First column – Candidate objects, Second column – Coordinate values for boundary of candidate objects shown in first column, Third column – Reconstructed candidate objects using small set of Fourier descriptors

Radon Descriptor

The Radon transform is a projection of the image intensity along a radial line oriented at a specific angle. Radon transform is applied to the detected candidate objects region. In general, the Radon transform of an image gives the projection of the candidate objects from 1 degree to 180 degrees. The experimentally it is observed that the Radon transform employed on candidate object region with a projection angle 30 degrees give accurate representation of the detected objects shape. The Radon transform is scale and rotation invariant, which can be observed in the Figure 6 and Figure 7 respectively, where the behavior of Radon Transform is constant for variation in rotation and scaling of candidate objects.

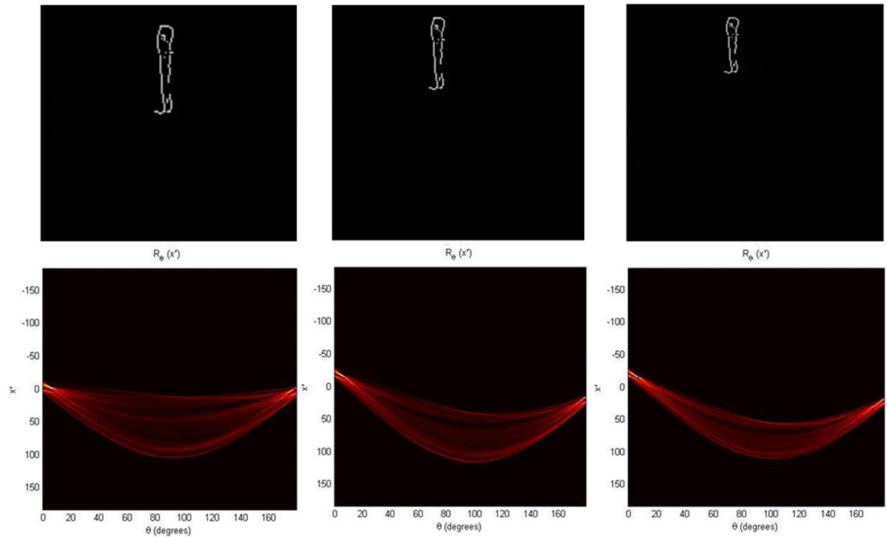

Figure 6: Behavior of Radon transform for scaling:  First row - candidate object with different scale, Second row – Corresponding result of Radon transform.

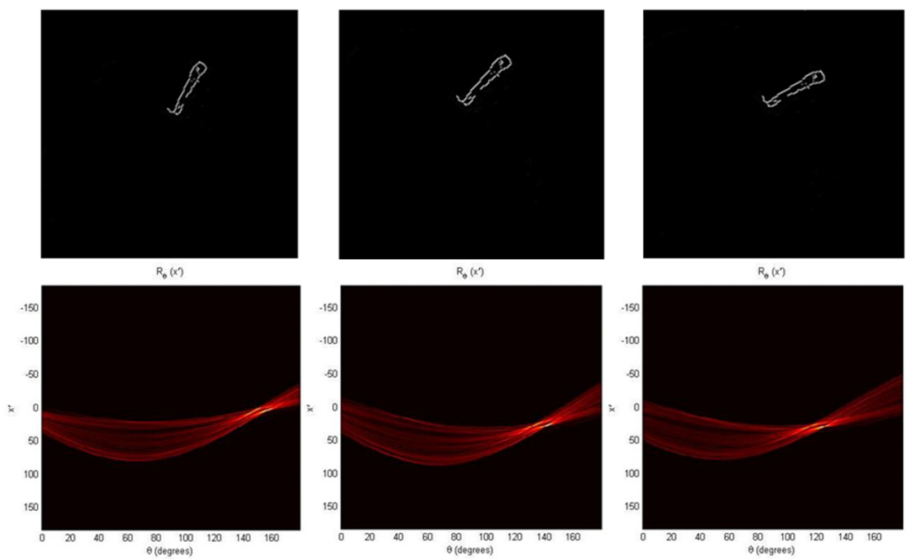

Figure 7: Behavior of Radon transform for orientation:  First row - candidate object with different orientation, Second row – Corresponding result of Radon transform.

### 3.3 Stem - Calyx Recognition

In the training phase, we extracted the shape features of detected stem/calyx and defected parts using three descriptors. The descriptors values or shape feature values are analyzed in order to identify the discriminating power. It is observed that the extracted shape features of stem-calyx parts are differing in nature compared to defect parts features. This motivated us to use the combination of above mentioned descriptors to differentiate stem-calyx parts from true defects, which increases the discrimination rate of stem-calyx from true defects. In order to differentiate the stem-calyx from true defects, classifier is employed. We have conducted experiments to find out suitable classifier, which yields highest recognition accuracy. We considered popular supervised classification algorithms such as Support Vector Machine (SVM) (Vladimir N. Vapnik, 1995), AdaBoost (Yoav Freund and Robert Schapire, 1995), Artificial Neural Network (ANN) (Robert Hecht-Nielsen, 1986), Linear Discriminant Classifier (LDC) (R. A. Fisher, 1936) and k-Nearest Neighbor classifier (k-NN) (Fix and Hodges, 1967).

## 4. EXPERIMENTAL RESULTS

We have conducted the experiments to evaluate the proposed method for recognizing stem-calyx part and differentiated from true defects of an apple. Since, there is no benchmark database of apples is available, we have created a Jonagold apple database for experimentation purpose. Image acquisition device is composed of a resolution 2.0 mega pixel (1240 x 1600). The camera is capable of acquiring only one-view of an apple. The healthy and small defected Jonagold apples were selected for experimentation purpose based on visual inspection. The images of the selected 200 Jonagold apples were captured in indoor environment with variation in lighting condition. Each apple was captured in three views such as stem part in view, calyx part in view and defected part in view.

The database contains 600 sample images of each image size 256 x 256 and 200 apple image samples for each view. Among the above-said data samples, we selected randomly 75% (from each view 150 samples) of the apple samples for training purpose, and the remaining 25% (from each view 50) samples are used for testing purpose. In order to eliminate the intersection (duplication) of training and testing sets, drop-one-out approach is used in this work. That is, an apple used for training is excluded from the testing set. In order to find out the optimal parameter for classifier, we have conducted several experiments using training and test sample. We considered only multifractal dimension descriptor to represent stem-calyx and defected parts. After conducting several experiments, it is observed that for SVM, polynomial kernel with a degree 3 yields highest recognition rate. A two-layer ANN is used to recognize stem-calyx. It is a feed-forward, error back-propagation network with adaptive learning rate. Several models of ANN is tested, however the one with 10 hidden nodes perform better than others. 3 inputs, 10 hidden with 2 output nodes are used and similarly k=4 for k-NN classifier. In the following experiments, classification algorithms with these best parameters are used.

Number of Fourier Descriptors

 The experiments are conducted to identify the small subset of Fourier descriptors required to represent overall features of the stem, calyx and defects. In these experiments, we used only Fourier descriptor to represent the stem, calyx and defects and k-NN classifier is used for classification. The Figure 8 shows that a small number of Fourier descriptors are sufficient for shape representation of stem, calyx and defected part. In our experiments, as can be seen in Figure 8, using more than 40 descriptors generally does not increase the recognition rate any further for defected parts. The shape of the stem, calyx and defected parts may undergo a lot of variation such as scaling and rotation. Therefore, in order to reflect such changes, we have limited the number of descriptors to 95 for stem, 130 for calyx and 40 for defected parts in all the experiments.

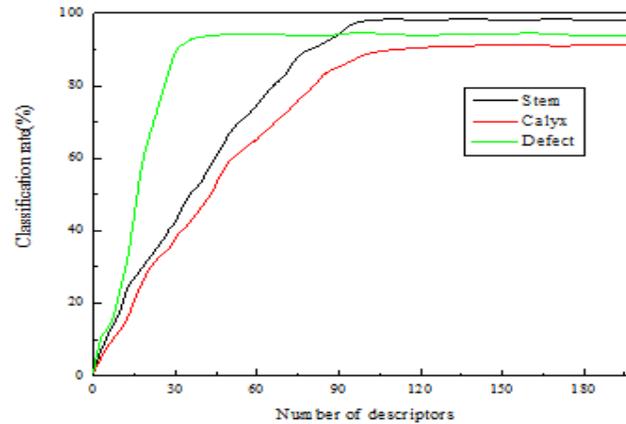

Figure 8. The Classification rate graph of the Fourier descriptors for stem, calyx and defects

Fusion of descriptors

The literature survey reveals that fusion of descriptors yields highest classification accuracy compared to an individual descriptor alone. We selected feature concatenation as the fusion approach. Three feature vectors $m_i$ (Multifractal descriptor), $f_i$ (Fourier descriptor) and $r_i$ (Radon descriptor) are concatenated into a single feature vector $v_i = (m_i, f_i, r_i)$ that is used for classification. The distance in the concatenated feature vector is defined as

$$d_t (v_i, v_j) = d(m_i, m_j) + d(f_i, f_j) + d(r_i, r_j). \qquad (11)$$

In order to verify the effectiveness of fusion of descriptors, we conducted experiments on fusion of descriptors. The Table 1 shows that there is an increase in the recognition rate when descriptors are fused. We fused the multifractal descriptor with Radon descriptor, which increases the recognition accuracy than multifractal descriptor alone. Similar effect was found for all the classifiers when we fused all three descriptors. Particularly, SVM classifier with the polynomial degree 3 yields highest recognition rate compared to other classifier when all three descriptors are combined and the results are shown in the Table 1. This concludes that the fusion of three descriptors helps to improve the classification rate compared to individual descriptor alone. Figure 9 describes, different levels of training datasets are used and classification rate is improved with increase in training dataset.

Table 1. Demonstration of increase in recognition rate (%) for fusion of shape descriptors

| Descriptors / Classifiers | MD | MD + RD | MD + RD + FD |
|---|---|---|---|
| SVM (Polynomial degree 3) | 92.40 | 93.60 | 96.00 |
| LDC | 91.20 | 92.40 | 94.80 |
| AdaBoost | 90.80 | 91.60 | 93.20 |
| ANN | 90.40 | 91.20 | 92.80 |
| k-NN (k=4) | 88.40 | 89.60 | 91.60 |

(MD – Multifractal Descriptor, RD – Radon Descriptor, FD – Fourier Descriptor)

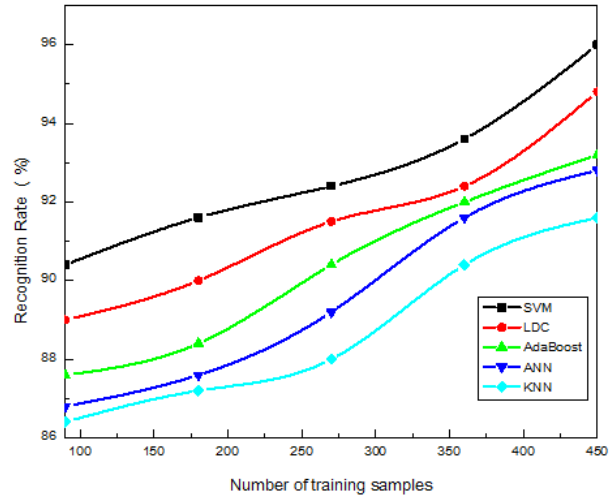

Figure 9. Classification accuracy of different classifiers for varying training samples
Evaluation of results

Table 2 shows the evaluation of experimental results using True Positive Rate (TPR) and False Positive Rate (FPR) obtained using SVM classifier with polynomial degree 3. We used 50 samples for each view such as stem, calyx, defects, stem with small defects and calyx with small defects. The TPR indicates the correct classification i.e., stem-calyx is classified as stem-calyx and defect is classified as defect. The FPR indicates the false classification i.e, stem-calyx is classified as defect and defect is classified as stem-calyx. From the results shown in the Table 2 indicates that proposed shape descriptors have shown high discrimination between stem-calyx and defects. The TPR for stem is approximately near to 100% classification accuracy and only 2% of classification error. This is not true in case of calyx recognition, only 47 calyx regions were correctly classified and remaining 03 regions were misclassified. This is due to the fact that shape features of small defects and calyx appear to be similar. Therefore, the classifier failed to distinguish these regions. Out of 100 test samples of stem and calyx regions, 04 regions were misclassified and classification accuracy of our method is 96% for Jonagold apples. Out of 50 defects samples, only 01 defect is misclassified as stem-calyx. The literature survey reveals that most of the stem-calyx recognition techniques uses apple with either stem-calyx or defects view. There may be possibility that both stem-calyx and defects lie on the same view. Therefore, our testing sample set consists of apple image including both stem-calyx and defects. The classification accuracy for this test set is promising. These rates show significant improvement with respect to the stem-calyx recognition results available in the literature.

Table 2. TPR and FPR for stem-calyx classification using SVM classifier

| **Different views of an apple** | **TPR** | **FPR** |
|---|---|---|
| Stems (n=50) | 49 (98%) | 01 |
| Calyxes (n=50) | 47 (94%) | 03 |
| Defects (n=50) | 49 (98%) | 01 |
| Stem with small defects (n=50) | 48 (96%) | 02 |
| Calyx with small defects (n=50) | 47 (94%) | 03 |

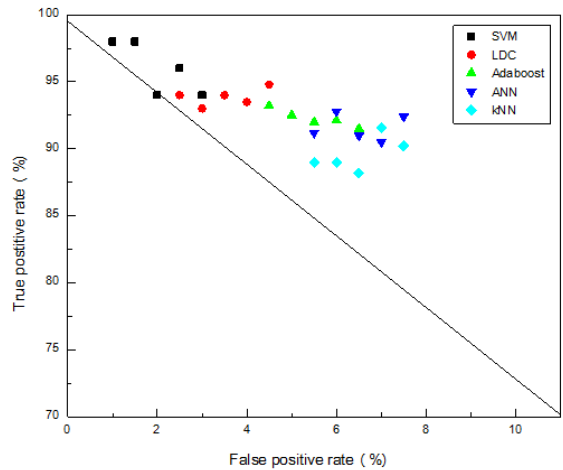

Figure 10. Classification performances of classifiers using TPR and FPR

The Figure 10 displays the performance of the classifiers using the results of True Positive Rate (TPR) and False Positive Rate (FPR). The diagonal line indicates the separation between the perfect recognition and perfect rejection, because in the graph left-top most part is the best part i.e., the best classification rate yields under this location and it presents the highest TPR and low FPR. Here, SVM yields the high TPR and low FPR among the classifiers, so we can easily say that SVM is the best classifier for our technique.

Discussion

The results of pre-processing showed that detection of candidate objects such as stem-calyx and small defects is visually true. A promising observation is that none of the stem-calyx and defects in the images is missed by proposed object segmentation step. The Jonagold was chosen for study because its mixed colors on the surface made it a more challenging and hence convincing example to demonstrate our algorithms performance. Other varieties such as red delicious, golden delicious and granny smith that other studies used in the past, all have uniform colors on the surface, which makes stem-calyx and defects detection an easier task.

The experimental results on stem-calyx recognition showed that proposed method yields highest classification accuracy for classification of stem-calyx regions and distinguished from true defects with small error rate for the Jonagold apple images captured in different orientation and scaling. The final classification rate of about 98% for stem ends and 94% for calyxes were achieved on the selected test samples. The average classification accuracy for both defective apples with stem end view and defective apples with calyx end view were 95%. The increase in the classification accuracy is due to the fact that discriminating shape features are extracted using scale and rotation invariant feature descriptors.  We observed that fusion of descriptors improves the classification rate compared to individual shape descriptors. The proposed algorithm failed to distinguish stem-calyx from true defects when, i) the apple images are captured in very high specular reflection (non-uniformity of color), ii) the stem end or calyx appeared near the edge of the observed apple surface, iii) small defects appeared very near the stem-ends or calyxes and iv) the apples are having many defects with the big size.

## 5. CONCLUSION

In this paper, we proposed an approach to recognize the stem-calyx of Jonagold apples and distinguished from the true defects. The grow-cut and multi-threshold techniques results are combined to detect the candidate objects. We extracted the shape features from detected candidate objects using three shape descriptors such as Multifractal, Fourier and Radon descriptors. The extracted shape features are concatenated and used to discriminate stem-calyx from true defects. We tested different classifiers such as SVM, ANN, k-NN, LDC and AdaBoost to discriminate stem-calyx from true defects. The SVM classifier with polynomial kernel function yields highest best recognition rate. The proposed method can be incorporated in apple grading technique as one of the steps in order to increase the apple grading accuracy.

## ACKNOWLEDGEMENTS

We would like to thank the anonymous reviewers for their critical comments and suggestions, which helped to improve the previous version of this paper.